%% file: TCMPI.tex
\documentclass[journal,transmag]{IEEEtran}
\usepackage{setspace,amssymb,amsmath,cite,enumerate,graphicx,multirow,subfigure,xcolor}
\usepackage[linesnumbered,ruled,vlined]{algorithm2e}
\newcommand{\tabincell}[2]{\begin{tabular}{@{}#1@{}}#2\end{tabular}}
\begin{document}

\title{Intelligent Optimization of Diversified Community Prevention of COVID-19 using Traditional Chinese Medicine}

\author{\IEEEauthorblockN{Yu-Jun~Zheng\IEEEauthorrefmark{1},
Si-Lan Yu\IEEEauthorrefmark{1},
Jun-Chao Yang\IEEEauthorrefmark{2},
Tie-Er Gan\IEEEauthorrefmark{2},
Qin Song\IEEEauthorrefmark{3},
Jun Yang\IEEEauthorrefmark{3},
and M\"{u}mtaz Karata\c{s}\IEEEauthorrefmark{4}}
\IEEEauthorblockA{\IEEEauthorrefmark{1}School of Information Science and Engineering, Hangzhou Normal University, Hangzhou 311121, China}
\IEEEauthorblockA{\IEEEauthorrefmark{2}First Affiliated Hospital, Zhejiang Chinese Medical University, Hangzhou 310006, China}
\IEEEauthorblockA{\IEEEauthorrefmark{3}Medical College, Hangzhou Normal University, Hangzhou 311121, China}
\IEEEauthorblockA{\IEEEauthorrefmark{4}Industrial Engineering Department, Naval Academy, National Defense University, Istanbul 34940, Turkey}% <-this % stops an unwanted space
%\thanks{Manuscript received June 21, 2020; revised .
%Corresponding author: Y.J. Zheng (email: yujun.zheng@computer.org).}
}

%\markboth{IEEE COMPUTATIONAL INTELLIGENCE MAGAZINE}{Zheng \MakeLowercase{\textit{et al.}}: Intelligent Optimization of COVID-19 Prevention with TCM}

\IEEEtitleabstractindextext{\begin{abstract}
Traditional Chinese medicine (TCM) has played an important role in the prevention and control of the novel coronavirus pneumonia (COVID-19), and community prevention has become the most essential part in reducing the spread risk and protecting populations. However, most communities use a uniform TCM prevention program for all residents, which violates the ``treatment based on syndrome differentiation'' principle of TCM and limits the effectiveness of prevention. In this paper, we propose an intelligent optimization method to develop diversified TCM prevention programs for community residents. First, we use a fuzzy clustering method to divide the population based on both modern medicine and TCM health characteristics; we then use an interactive optimization method, in which TCM experts develop different TCM prevention programs for different clusters, and a heuristic algorithm is used to optimize the programs under the resource constraints. We demonstrate the computational efficiency of the proposed method and report its successful application to TCM-based prevention of COVID-19 in 12 communities in Zhejiang province, China, during the peak of the pandemic.
\end{abstract}

}

\maketitle
%\IEEEdisplaynontitleabstractindextext
\IEEEpeerreviewmaketitle

\section{Introduction}
\IEEEPARstart{T}he ongoing outbreak of the novel coronavirus pneumonia (COVID-19), declared by the World Health Organization as a global public health emergency, has been reported in over ten million cases in over 200 countries and territories as of June 29, 2020. Community prevention and control has become the most basic and essential part in reducing the spread risk and protecting populations during the pandemic \cite{Wu20JAMA,Zhang20AJIC}. Currently, community prevention is a significant challenge, not only because there is still no effective antiviral or vaccine, but also because of the pressing need to restart economy and restore social life \cite{McKee20NatMed,Zheng20ORT}.

Although modern medicine offers accurate diagnosis and treatment methods for many diseases, it shows weakness in preventing emerging infectious diseases such as COVID-19 for which there is no vaccine, because epidemic prevention solutions based on modern medicine heavily rely on a clear understanding of the pathogenic mechanism and a number of large case-controlled studies \cite{Lore14CMLS,Feng20CJEpi}, and the misuse of antibiotics can cause severe side effects \cite{Smith05PNAS}.

Traditional Chinese medicine (TCM) has been developed and used in the prevention and treatment of various diseases for thousands of years in Chinese history. TCM is a comprehensive system of the treatment of acute and chronic disorders as well as for the prevention of such disorders mainly based on herb medicine \cite{WangX12AJCM}. Unlike modern medicine that focuses on killing viruses, TCM pays attention to improving the inherent self-resistance and reducing the likelihood of disease onset by using a unique holistic approach to establish equilibrium in the whole and individual parts of the body \cite{Zhang12CTMed}. The present principles on prevention of COVID-19 are to tonify body energy to protect the outside body, dispel wind, dissipate heat and dissipate dampness \cite{Wang20IJAntiAge}. Facing an emerging infectious disease, TCM prescriptions (formulae) are made by combining existing crude herbs or minerals instead of developing new drugs. That is why TCM have achieved great success in response to recent epidemics such as SARS, H1N1, and Zika \cite{Liu04JACMed,Leung07AJCM,Ji10CJCMM,Chang11PLosCB,Lu16APCT}, and is playing a vital role in reducing the incidence rate and controlling the spread of COVID-19 \cite{LuoH20CJIMed,Ren20PharRes,XuJ20CTCP}.

However, we believe that there is much room for improvement of community prevention of the pandemic using TCM. For example, since the COVID-19 outbreak in China, many local public health administrations have issued TCM prevention programs for COVID-19, and some communities used a single program or prescription for all residents \cite{Chan20AJCMed,Ang20IMedRes}. Such a one-size-fits-all solution violates the ``treatment according to three factors (time, place, and people)ÈýÒòÖÆÒË'' and ``treatment based on syndrome differentiation'' principles of TCM and, therefore, limits the effectiveness of prevention.

According to requirements of local governments to improve community prevention of COVID-19, we propose an intelligent, diversified community prevention method for COVID-19 by combining TCM and modern computational intelligence methods. First, we use a fuzzy clustering method to divide the population based on both modern medicine and TCM health characteristics. According to the health characteristics, TCM experts develop a TCM prevention program for each cluster. The initial program for each cluster aims to maximize the prevention effect on residents in the cluster. Nevertheless, the demands of all initial programs often exceed the available resources. We then use an interactive optimization method to continually evolve the prevention programs until all programs are approved by the TCM experts while all resource demands are satisfied. The proposed method has been successfully practiced in a number of communities in Zhejiang province and extended to many other regions in China during the peak of COVID-19.

%The remainder of this paper is organized as follows. Section \ref{sec:prob} presents the medical supplies procurement problem. Section \ref{sec:oldalg} simply describes how to directly use basic multiobjective EAs to solve the original problem. Section \ref{sec:newalg} proposes the transform-and-divide evolutionary optimization approach. section \ref{sec:exp} presents the computational results. Section \ref{sec:conclu} concludes with a discussion.

\section{Fuzzy Clustering of Community Residents}
Using a ``one-size-fits-all'' prevention program for all community residents fails to consider interpersonal differences and is rarely effective. On the contrary, developing a personalized prevention program for each resident would be too expensive due to the limited resources under the pandemic.

We prefer to develop a set of diversified prevention programs, each for a group of residents with similar physical characteristics. The characteristics include both modern medicine characteristics and TCM health characteristics, as summarized in Table \ref{tab1}. The data sources of the characteristics include both modern healthcare records and TCM records of the residents. Those characteristics are used as input features for grouping, and the value of each feature is normalized, e.g., real values (such as height and weight) are normalized to [0,1], and exclusive labels (such as sex and illness) are represented by binary variables. In this study, we use a total of 1148 features for grouping. However, for most residents, a large portion of features are inevitably absent. For example, a resident typically is free of illness or with only one illness.

\begin{table*}
\center\footnotesize%\setlength\tabcolsep{5pt}\renewcommand{\arraystretch}{0.9}
\caption{Physical characteristics for grouping community residents.}
\begin{tabular}{l|lcc}\hline
\multicolumn{1}{c|}{Type} &\multicolumn{1}{c}{Characteristics} &\multicolumn{1}{c}{Number of indicators} \\\hline
Basic health metrics &age, sex, occupation, height, weight, heart rate, blood pressure, vital capacity, \ldots &121\\
TCM constitutions & \tabincell{l}{mild, yang deficiency, yin deficiency, phlegm dampness, wet \& heat, qi stagnation,\\ qi deficiency, blood stasis, special}& 9\\
TCM syndromes \cite{Liu11TCMSyn} &\tabincell{l}{shire jinyin, pixu shiyun, xuexu fengzao, shire yuzu, shire shangyin, qizhi xueyu, \\qixu buzu, ganshen buzu,  \ldots} & 50\\
Past illnesses & diseases and the corresponding indicators &484\\
Current illnesses & diseases and the corresponding indicators &484\\\hline
\end{tabular}
\label{tab1}\end{table*}

\subsection{An Improved Fuzzy $c$-Means Clustering Method}
For such a high-dimension grouping problem with many missing values, classical hard clustering methods such as $K$-means clustering\cite{Kanu02PAMI} are not very effective. In this study, we use an improved fuzzy $c$-means (FCM) clustering method \cite{Bezdek81}. In brief, FCM groups a set of data points by minimizing the overall fuzzy-membership-weighted distance of the data points from cluster centroids:
\begin{equation}
\min J(U,V)= \frac{1}{cn}\sum_{i=1}^{c}\sum_{j=1}^{n}u^m_{ij}d^2_{ij}
\label{eq:fcm}\end{equation}
where $n$ is the number of data points, $c$ is the number of clusters, $u_{ij}$ is the membership degree of the $j$th data point to the $i$th cluster subject to $\sum_{i}u_{ij}\!=\!1$, $d_{ij}$ is the distance between the $j$th data point and the $i$th cluster centroid, $m$ is a control parameter with a default value of 2, $U=(u_{ij})_{c\!\times\!n}$ is the weight matrix, and $V\!=\![\mathbf{v}_1,\mathbf{v}_2,\ldots,\mathbf{v}_c]$ is the set of cluster centroid vectors.

Xu and Wu \cite{Xu10JSEE} extended the standard FCM to intuitionistic FCM based on new distance measures defined on intuitionistic fuzzy sets (IFS) \cite{Atan94FSS} that capture more uncertainty information. Here, we use Pythagorean fuzzy sets (PFS) \cite{Yager13NAFIPS}, which allow for a larger body of membership grades than IFS, to further improve clustering. Formally, let $S$ be an arbitrary non-empty set, a PFS is a mathematical object of the form:
\begin{equation}
P= \{\langle x,P(\mu_P(x),\nu_P(x))\rangle|x\in S\}
\end{equation}
where $\mu_P(x):S\rightarrow[0,1]$ and $\nu_P(x):S\rightarrow[0,1]$ are respectively the membership degree and the non-membership degree of the element $x$ to $S$ in $P$. PFS extends IFS in that the membership degrees satisfy $\mu_P^2(x)+\nu_P^2(x)\le1$. The hesitant degree of $x\in X$ is expressed as:
\begin{equation}
\pi_P(x)=\sqrt{1-\mu_P^2(x)-\nu_P^2(x)}
\end{equation}

In our improved Pythagorean FCM, data points and cluster centroids are represented by Pythagorean fuzzy number vectors, where the distance between two Pythagorean fuzzy numbers $\beta_1\!=\!P(\mu_{\beta_1},\nu_{\beta_1})$ and $\beta_2\!=\!P(\mu_{\beta_2},\nu_{\beta_2})$ is calculated as \cite{Ren16ASOC}:
\begin{equation}
|\beta_1, \beta_2|= \sqrt{\frac{(\mu^2_{\beta_1}\!-\!\mu^2_{\beta_2})^2+ (\nu^2_{\beta_1}\!-\!\nu^2_{\beta_2})^2+ (\pi^2_{\beta_1}\!-\!\pi^2_{\beta_2})^2} {2}}
\end{equation}
%where $\pi_{\beta}=\sqrt{1\!-\!\mu^2_{\beta}\!-\!\nu^2_{\beta}}$ is the hesitant degree of $\beta$.

And the distance $d_{ij}$ in Eq. (\ref{eq:fcm}) between a $D$-dimensional data point $\mathbf{x}_j$ and a cluster centroid $\mathbf{v}_i$ is calculated as:
\begin{equation}
\|\mathbf{x}_j,\mathbf{v}_i\|= \sqrt{\frac{\sum_{d=1}^D |x_{jd}, v_{id}|^2} {D}}
\end{equation}

Algorithm \ref{alg:pfcm} presents the Pythagorean FCM method, which minimizes the objective function $J(U,V)$ by iteratively updating the fuzzy membership weights (Line 11) and centroids (Line 13) to apply the derivative of $J(U,V)$ \cite{Xu10JSEE}.

\begin{algorithm}
\small
Initialize a $c\!\times\!n$ matrix $U$ and a set $V^{(0)}$ of $c$ cluster centroids\;
Let $k\!=\!0$\;
\While{$\|V^{(k\!+\!1)},V^{(k)}\|>\epsilon$}{
    \For{$j=1$ to $n$}{
        \If{$\exists i':1\!\le\!i'\!\le\!c: \|\mathbf{x}_j,\mathbf{v}_{i'}\|=0$}{
            \For{$i=1$ to $c$}{
                \lIf{$i=i'$}{$u^{(k)}_{ij}\gets 1$\;}
                \lElse{$u^{(k)}_{ij}\gets 0$\;}
            }
        }
        \Else{
            \For{$i=1$ to $c$}{
                $u^{(k)}_{ij}\gets \frac{1}{\sum\limits_{i'\!=\!1}^c \big(\frac{\|\mathbf{x}_j,\mathbf{v}_i\|}{\|\mathbf{x}_j,\mathbf{v}_{i'}\|}\big) ^{\frac{2}{m\!-\!1}}}$\;
            }
        }
    }
    \For{$i=1$ to $c$}{
        $\mathbf{v}^{(k)}_i\gets \left\{\left\langle \beta_d, \frac{\sum_{j\!=\!1}^n u^{(k)}_{ij}\mu^2_{\beta_d}}{\sum_{j\!=\!1}^n u^{(k)}_{ij}}, \frac{\sum_{j\!=\!1}^n u^{(k)}_{ij}\nu^2_{\beta_d}}{\sum_{j\!=\!1}^n u^{(k)}_{ij}} \right\rangle| 1\!\le\!d\!\le\!D\right\}$\;
    }
    $k\gets k\!+\!1$\;
}
\Return $(U,V)$\;
\caption{Pythagorean fuzzy $c$-means clustering algorithm.}
\label{alg:pfcm}\end{algorithm}

\subsection{A Metaheuristic for Optimizing Clusters}\label{sec:wwo-fcm}
The performance of FCM clustering heavily depends on the quality of initial cluster centroids \cite{Gong13TIP,Zhang19SOCO}. Instead of randomly setting initial cluster centroids, we use a metaheuristic algorithm, ecogeography-based optimization (EBO) \cite{Zheng14CAOR}, to optimize initial cluster centroids \cite{Zhang19SOCO}. The algorithm starts by initializing a population of solutions, each representing a set $V^{(0)}$ of $c$ initial cluster centroids. Let $J(\mathbf{x})$ denote the resulting $J(U,V)$ value obtained by the FCM method from the initial cluster centroids of $\mathbf{x}$; each solution $\mathbf{x}$ is assigned with an emigration rate $E_r(\mathbf{x})$ that is inversely proportional to $J(\mathbf{x})$ and an immigration rate $I_r(\mathbf{x})$ that is proportional to $J(\mathbf{x})$:
\begin{eqnarray}
E_r(\mathbf{x})&=& \frac{J_{\max}-J(\mathbf{x})+\epsilon}{J_{\max}-J_{\min}+\epsilon} \label{eq:emr}\\
I_r(\mathbf{x})&=& \frac{J(\mathbf{x})-J_{\min}+\epsilon}{J_{\max}-J_{\min}+\epsilon} \label{eq:imr}
\end{eqnarray}
where $J_{\max}$ and $J_{\min}$ are the maximum and minimum $J(\cdot)$ values among the population, and $\epsilon$ is a small positive number to avoid division-by-zero. In this way, a better solution has a higher probability of emigrating features to other solutions, while a worse solution has a higher probability of immigrating features from other solutions \cite{Simon08TEC}.

The EBO algorithm then continually evolves the solutions using two migration operators: local migration and global migration. Local migration updates a solution $\mathbf{x}$ at each dimension $d$ by migrating the corresponding dimension of a neighboring solution $\mathbf{x}^\dag$ as follows:
\begin{equation}
x'_d= x_d+\textit{rand}(0,1)\cdot(x^\dag_d-x_d)
\label{eq:lmig}\end{equation}
where $\textit{rand}(0,1)$ produces a random number uniformly distributed in $(0,1)$, and $\mathbf{x}^\dag$ is selected from all neighbors of $\mathbf{x}$ with a probability proportional to $E_r(\mathbf{x}^\dag)$.

Global migration updates a solution $\mathbf{x}$ at each dimension $d$ by migrating the corresponding dimensions of both a neighboring solution $\mathbf{x}^\dag$ and a non-neighboring solution $\mathbf{x}^\ddagger$ as follows:
\begin{equation}
x'_d= \left\{\begin{array}{ll}
    x^\dag_d+\textit{rand}(0,1)\cdot(x^\ddagger_d-x_d), & f(\mathbf{x}^\ddagger)\!\le\! f(\mathbf{x}^\dag)\\
    x^\ddagger_d+\textit{rand}(0,1)\cdot(x^\dag_d-x_d), & f(\mathbf{x}^\ddagger)\!>\! f(\mathbf{x}^\dag)
\end{array}\right.
\label{eq:gmig}\end{equation}
where $\mathbf{x}^\dag$ is selected from all neighbors of $\mathbf{x}$ with a probability proportional to $E_r(\mathbf{x}^\dag)$, and $\mathbf{x}^\ddagger$ is selected from all other solutions that are not neighbors of $\mathbf{x}$ with a probability proportional to $E_r(\mathbf{x}^\ddagger)$.

EBO uses a parameter $\eta$ as the probability of performing global migration and, therefore, ($1\!-\!\eta$) as the probability of performing local migration. The value of $\eta$ dynamically increases from a lower limit $\eta_{\min}$ to an upper limit $\eta_{\max}$ with generation $g$ of the algorithm:
\begin{equation}
\eta= \eta_{\min}+ \frac{g}{g_{\max}}(\eta_{\max}\!-\!\eta_{\min})
\label{eq:eta}\end{equation}

In this study, we use a local random neighborhood structure \cite{Zheng14SOCO}, which randomly assigns $k_N$ neighboring solutions to each solution in the population (where $k_N$ is a control parameter); if the current best solution has not been updated after a number $\widehat{g}$ of consecutive generations, the neighborhood structure is randomly reset. Algorithm \ref{alg:ebo} presents the pseudocode the EBO algorithm, where Line 4 invokes Algorithm 1 to evaluate the fitness of each solution (initial centroid setting).

\begin{algorithm}\small
Randomly initialize a population of solutions (initial set of cluster centroids)\;
\While{the stopping criterion is not satisfied}{
    \ForEach{solution $\mathbf{x}$ in the population}{
        Use Algorithm \ref{alg:pfcm} to produce the clustering results $(U,V)$ from the initial cluster centroids of $\mathbf{x}$\;
    }
    Let $\mathbf{x}^*$ be the best solution in the population\;
    \ForEach{solution $\mathbf{x}$ in the population}{
        Compute $E_r(\mathbf{x})$ and $I_r(\mathbf{x})$ according to Eqs. \eqref{eq:emr} and \eqref{eq:imr}\;
    }
    \ForEach{solution $\mathbf{x}$ in the population}{
        \For{$d=1$ to $n$}{
            \If{$\textit{rand}(0,1)<I_r(\mathbf{x})$}{
                Select a neighboring $\mathbf{x}^\dag$ with probability proportional to $E_r(\mathbf{x}^\dag)$\;
                \If{$\textit{rand}(0,1)<\eta$} {
                    Select a non-neighboring $\mathbf{x}^\ddag$ with probability proportional to $E_r(\mathbf{x}^\ddag)$\;
                    Perform global migration according to Eq. (\ref{eq:gmig})\;
                }
                \Else{
                    Perform local migration according to Eq. (\ref{eq:lmig})\;
                }
            }
        }
        \If{the migrated solution $\mathbf{x}'$ is better than $\mathbf{x}$}{
            $\mathbf{x}\gets\mathbf{x}'$\;
        }
    }
    Update $\eta$ according to Eq. (\ref{eq:eta})\;
    \If{$\mathbf{x}^*$ has not been updated for $\widehat{g}$ consecutive generations}{
        Randomly reset neighborhood structure\;
    }
}
\Return the clustering result of the best known solution $\mathbf{x}^*$.
\caption{The EBO algorithm for enhancing the fuzzy clustering method.}
\label{alg:ebo}\end{algorithm}

\section{Interactive Optimization of Prevention Programs}\label{sec:prob}
After clustering the community residents into $c$ groups, we invite TCM experts to assess the health characteristics of each cluster by examining representative residents, and develop diversified prevention programs according to the characteristics. Note that the number $p$ of prevention programs approximates, but does not necessarily equal, the number $c$ of clusters. That is, the TCM experts typically develop a prevention program (including a TCM prescription and other supplementary measures such as acupuncture and moxibustion) for a cluster; they may also develop a prevention program for two similar clusters as they consider appropriate; for a cluster with high-risk residents (with suspected symptoms of COVID-19 or serious underlying illnesses), they can decide to develop one prevention program for each resident.

\begin{figure}
\centering\includegraphics[scale=0.75]{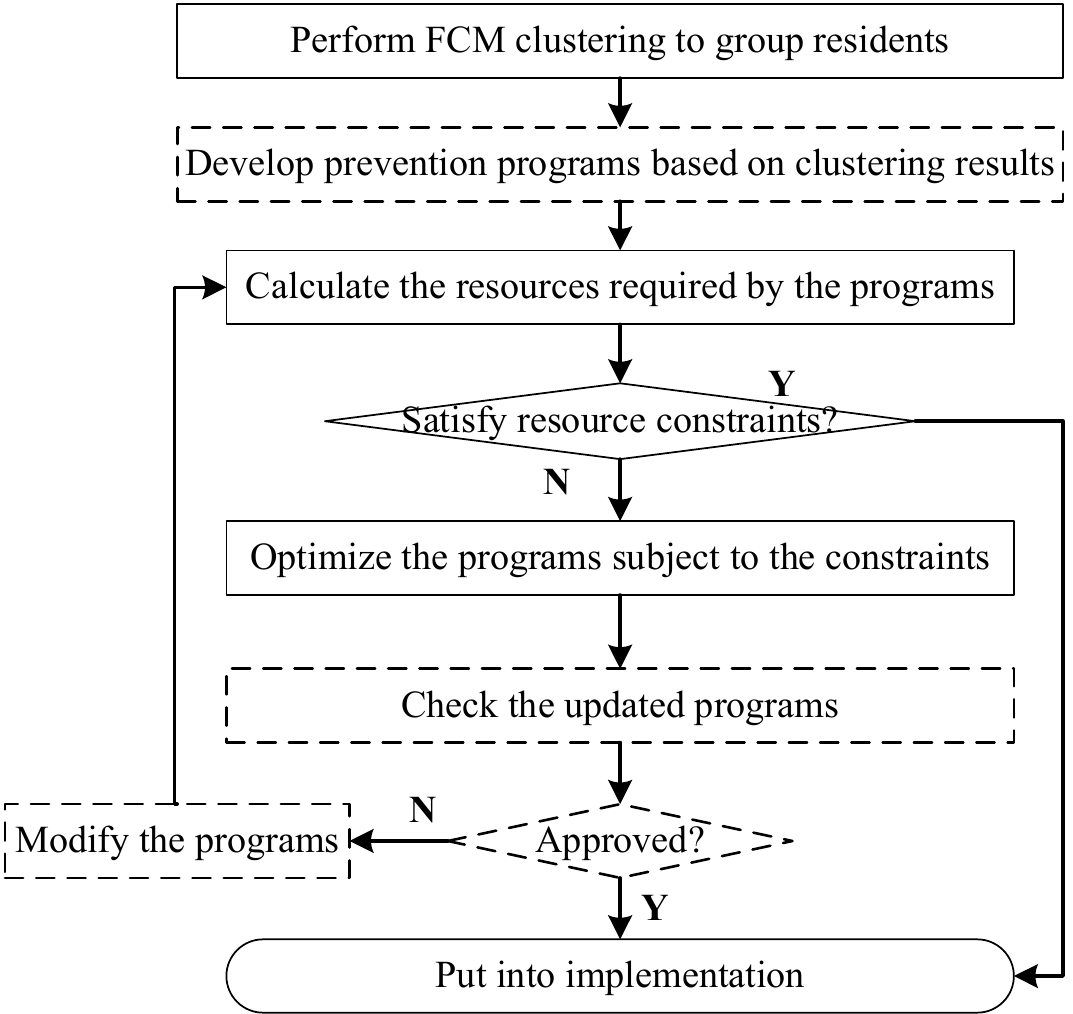}
\caption{Flowchart of the interactive optimization of TCM prevention programs. The rectangles with dash borders are performed by TCM experts, while the rectangles with solid borders are performed by computer.}
\label{fig:flow}\end{figure}

When developing initial programs, the TCM experts aim to maximize the prevention effect on residents of each cluster without considering the limits of medical resources, including herbal medicines, patent medicines, medical devices, pharmacists, and other paramedical personnel. If the demands of the programs exceed the available resources, we use an intelligent optimization algorithm to optimize the programs subject to the resource constraints. However, any prevention program produced by computer algorithms must be checked and, if necessary, modified by the TCM experts before implementation. The above process continues until all prevention programs satisfy the resource constraints and are approved by the TCM experts. Fig. \ref{fig:flow} illustrates the interactive optimization process.

\subsection{Optimization Problem}
The problem of optimizing prevention programs is formulated as follows. TCM experts have developed $N$ basic prevention programs, denoted by $\{P_1,P_2,\ldots,P_N\}$, for residents from $M$ communities, denoted by $\{C_1,C_2,\ldots,C_M\}$. The programs involve:
\begin{itemize}
\item $K$ types of drugs denoted by $\{D_1,D_2,\ldots,D_K\}$;
\item $K_1$ types of other medical resources that can be shared among the communities, denoted by $\{G_1,G_2,\ldots,G_{K_1}\}$, such as TCM material and movable devices;
\item $K_2$ types of other medical resources that cannot be shared among the communities, denoted by $\{F_1,F_2,\ldots,F_{K_2}\}$, such as immovable devices and staffs belong to given communities.
\end{itemize}

The problem is to optimize the distribution of medical resources among the prevention programs for the $M$ communities. In the local region, the total available quantity of each drug $D_k$ is $\widehat{q}^\textnormal{D}_k$  ($1\!\le\!k\!\le\!K$), total available quantity of each other sharable resource $G_k$ is $\widehat{q}^\textnormal{G}_k$ ($1\!\le\!k\!\le\!K_1$), and available quantity of each non-sharable resource $F_k$ in community $C_i$ is $\widehat{q}^\textnormal{F}_{ik}$ ($1\!\le\!k\!\le\!K_2;1\!\le\!i\!\le\!M$).

Each prevention program $P_j$ has the following attributes ($1\!\le\!j\!\le\!N$):
\begin{itemize}
\item The number $n_j$ of residents using the program;
\item The set $\Theta_j$ of communities that have residents using the program; for each community $C_i\!\in\!\Theta_j$, the number of residents using the program is $n_{ij}$;
\item The set $\Phi_j$ of drugs used by the program; for each drug $D_k\in\Phi_j$, the quantity used per prescription is $q^\textnormal{D}_{jk}$;
\item The set $\Psi_j$ of other sharable resources used by the program; for each resource $G_k\in\Psi_j$, the quantity used per prescription is $q^\textnormal{G}_{jk}$;
\item The set $\Omega_j$ of other non-sharable resources used by the program; for each resource $F_k\in\Omega_j$, the quantity used per prescription is $q^\textnormal{F}_{jk}$;
\end{itemize}

In a TCM prescription, many ingredients have alternatives. We use $\Phi'_j\subset\Phi_j$ to denote the subset of drugs that have alternatives in $P_j$; for each drug $D_k\in\Phi'_j$, we use a list $\Lambda_k$ to store its alternative drugs in decreasing order of priority, which are determined by TCM drug properties and effects to the disease (COVID-19 belongs to pulmonary disease in TCM). We consider two types of updates on a TCM prescription:
\begin{itemize}
\item Replacing an auxiliary drug $D_k\!\in\!\Phi'_j$ with an alternative $D_{k'}\!\in\!\Lambda_k$, for example, replacing coix seed with winter melon seed;
\item Replacing a main drug $D_k\!\in\!\Phi'_j$ with an alternative $D_{k'}\!\in\!\Lambda_k$; however, according to compatibility of TCM, a main drug is related to one or more auxiliary drugs, and the corresponding auxiliary drugs should also be reapplied; an example is replacing ``astragalus membranaceus (main) + cinnamon (auxiliary)'' with ``codonopsis pilosula (main) + yam (auxiliary)''.
\end{itemize}

To avoid the updated prescriptions deviating too much from the original prescriptions developed by TCM experts, for each prevention program, we allow updating at most one drug (except auxiliary drugs related to a main updated drug) at each time.
For either of the above two types of updates, we use $P_j(x_j,x'_j)$ to denote the updated program, where $x_j$ denotes the index of the original drug $D_k$ in the prescription, and $x_j$ denotes the index of the alternative drug $D_{k'}$ in $\Lambda_k$. Therefore, for the set of original prevention programs $\{P_1,P_2,\ldots,P_N\}$, each solution to the problem can be represented by a $(2N)$-dimensional integer vector $\mathbf{x}=\{x_1,x'_1,x_2,x'_2,\ldots,x_N,x'_N\}$, which indicates that the $x_j$-th drug in $P_j$ is to be replaced by its $x'_j$-th alternative ($1\!\le\!j\!\le\! N$); without loss of generality, $x_j\!=\!x'_j$ denotes that $P_j$ is unchanged.

Based on the efficacy of the original and alternative drugs, we can determine the quantity of an alternative drug $D_{k'}$ used to replace an original drug $D_k$ in the prescription. Based on the change of the prescription, we can then determine the changes of other medical resources, such as the types and quantities of material and the working hours for processing the drugs. Consequently, we obtain the following attributes of the updated prevention program $P_j(x_j,x'_j)$:
\begin{itemize}
\item The set $\Phi_j(x_j,x'_j)$ of drugs; for each drug $D_k\in\Phi_j(x_j,x'_j)$, the quantity used per prescription is $q^\textnormal{D}_{jk}(x_j,x'_j)$;
\item The set $\Psi_j(x_j,x'_j)$ of other sharable resources used by the program; for each resource $G_k\in\Psi_j(x_j,x'_j)$, the quantity used per prescription is $q^\textnormal{G}_{jk}(x_j,x'_j)$;
\item The set $\Omega_j(x_j,x'_j)$ of other non-sharable resources used by the program; for each resource $F_k\in\Omega_j(x_j,x'_j)$, the quantity used per prescription is $q^\textnormal{F}_{jk}(x_j,x'_j)$.
\end{itemize}

The objective of the problem is to maximize the overall effects of the updated prevention programs, subject to that the resources used by the programs do not exceed the available resources. The effect of each updated program $P_j(x_j,x'_j)$ is evaluated based on its deviation from the original program $P_j$: the larger the deviation, the smaller the effect is, as we should trust the ability of TCM experts who develop the original program. The deviation of $P_j(x_j,x'_j)$ from $P_j$ is assessed in two aspects:
\begin{itemize}
\item The importance of $D_{x_j}$ in the original $P_j$, which is measured by a weight $w_{jk}$; a larger priority indicates a larger deviation;
\item The priority of $D_{x'_j}$ in the alternative set $\Lambda_{x_j}$; a higher priority indicates a smaller deviation.
\end{itemize}

Here, we calculate the deviation as follows:
\begin{equation}
\Delta P_j(x_j,x'_j)= w_{jx_j} I(\Lambda_{x_j},x'_j)
\end{equation}
where $I(\Lambda_{x_j},x'_j)$ is the index of $D_{x'_j}$ in $\Lambda_{x_j}$ (without loss of generality, we set $\Delta P_j(x_j,x_j)\!=\!0$).

Moreover, we use a weight $w_j$ to denote the susceptibility of residents covered by program $P_j$ to the epidemic, and use a weight $w'_i$ to denote the importance of each community $C_i$ (which is related to the openness and population density of the community). The objective of the problem is defined as:
\begin{equation}
\min f(\mathbf{x})= \sum_{j=1}^N\sum_{C_i\in\Theta_j}w_jw'_i\Delta P_j(x_j,x'_j)
\label{eq:obj}\end{equation}

The constraints of the problem are the quantities of each drug, other sharable resource, and other sharable resource used by the programs cannot exceed available quantities:
\begin{eqnarray}
&& \sum_{j=1}^N n_jq^\textnormal{D}_{jk}(x_j,x'_j)\le \widehat{q}^\textnormal{D}_k, \quad 1\!\le\!k\!\le\!K \label{eq:conD}\\
&& \sum_{j=1}^N n_jq^\textnormal{G}_{jk}(x_j,x'_j)\le \widehat{q}^\textnormal{G}_k, \quad 1\!\le\!k\!\le\!K_1 \label{eq:conG}\\
&& \sum_{j=1}^N n_{ij}q^\textnormal{F}_{jk}(x_j,x'_j)\le \widehat{q}^\textnormal{F}_{ik}, \quad 1\!\le\!i\!\le\!M; 1\!\le\!k\!\le\!K_2 \label{eq:conF}
\end{eqnarray}

It should be noted that, in Eqs. (\ref{eq:conD})--(\ref{eq:conF}), we uniformly use the operator $\sum$ for notational  simplicity; however, it does not necessarily always be summation. Typically, for drugs and material, $\sum$ denotes summation; but for other resources such as devices and personnel, $\sum$ can be other corresponding aggregation operators. For example, suppose that a decocting machine can process 50 doses of a prescription, the operator will add 1 per 50 doses, and will also add 1 if the number of remaining doses is less than 50.

\subsection{Optimization Algorithm}\label{sec:newalg}
A TCM prescription can have tens of ingredients, and a drug can have tens of alternative drugs. Therefore, when the number $N$ of prevention programs is relatively large, the solution space of the problem can be very large, for which tradition exact optimization methods are often inefficient.

We use a metaheuristic optimization algorithm, water wave optimization (WWO) \cite{Zheng15COR}, to efficiently solve the problem. The algorithm starts by initializing a population of $N_P$ solutions. To evaluate the fitness of each solution $\mathbf{x}$, we employ three penalty functions $v_D(\mathbf{x})$, $v_G(\mathbf{x})$, and $v_F(\mathbf{x})$ to calculate the violations of constraints \eqref{eq:conD}, \eqref{eq:conG}, and \eqref{eq:conF} as follows:
\begin{eqnarray}
v_D(\mathbf{x})&\!=\!& \sum_{k=1}^K\max\big(0,\sum_{j=1}^N n_jn^\textnormal{D}_{jk}(x_j,x'_j)\!-\! \widehat{n}^\textnormal{D}_k\big) \label{eq:violD}\\
v_G(\mathbf{x})&\!=\!& \sum_{k=1}^{K_1}\max\big(0,\sum_{j=1}^N n_jn^\textnormal{G}_{jk}(x_j,x'_j)\!-\! \widehat{n}^\textnormal{G}_k\big) \label{eq:violG}\\
v_F(\mathbf{x})&\!=\!& \sum_{i=1}^M\sum_{k=1}^{K_2}\max\big(0,\sum_{j=1}^N n_{ij}n^\textnormal{F}_{jk}(x_j,x'_j)\!-\! \widehat{n}^\textnormal{F}_{ik}\big) \label{eq:violF}
\end{eqnarray}

And the fitness $\textit{fit}(\mathbf{x})$ is calculated as:
\begin{equation}
\textit{fit}(\mathbf{x})= 1/\big(f(\mathbf{x})+v_D(\mathbf{x})+v_G(\mathbf{x})+v_F(\mathbf{x})\big)
\label{eq:fit}\end{equation}

We sort all $N_P$ solutions in the population in decreasing order of the fitness value. Let $o(\mathbf{x})$ be the index of solution $\mathbf{x}$ in the sorted population, according to the principles of adapting WWO for combinatorial optimization \cite{Zheng19ASOC}, we calculate a wavelength $\lambda(\mathbf{x})$ for each $\mathbf{x}$ as an integer between 1 and $N$ as follows:
\begin{equation}
\lambda(\mathbf{x})= N-\bigg\lceil\frac{N_P-o(\mathbf{x})}{N_P-1}(N-1)\bigg\rceil
\label{eq:wavelength}\end{equation}
where $\lceil\cdot\rceil$ denotes rounding up to the nearest integer.

The WWO iteratively evolves the solutions using three operators including propagation, refraction, and breaking. The propagation operator is based on two neighborhood structures. Given a solution $\mathbf{x}$ to the problem, its neighboring solutions can be obtained using one of the following two approaches:
\begin{itemize}
\item Randomly selecting a prescription $P_j$, changing $x'_j$ to a random index of $\Lambda_{x_j}$; this indicates modifying the alternative drug used in $P_j(x_j,x'_j)$;
\item Randomly selecting a prescription $P_j$, changing $x_j$ to the index of another drug $D_k\in\Phi_j$, and then changing $x'_j$ to a random index of $\Lambda_k$; this indicates modifying both the drug to be replaced in $P_j$ and the alternative drug used in $P_j(x_j,x'_j)$.
\end{itemize}

\begin{figure}
\centering
\includegraphics[scale=0.8]{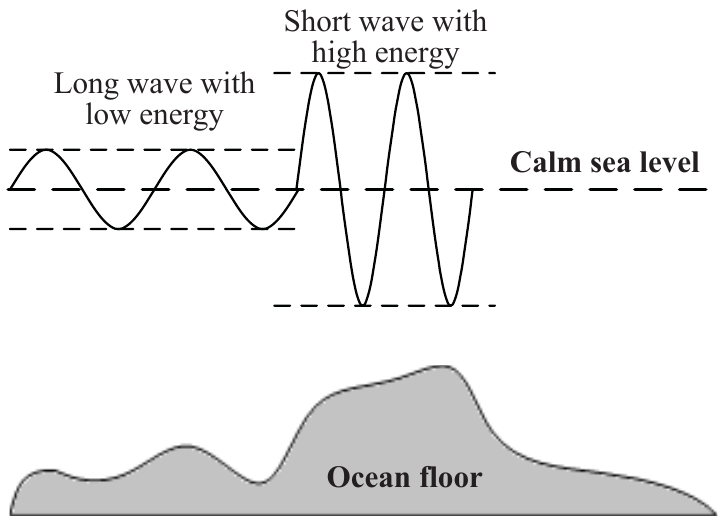}
\caption{Different wave lengths of high-fitness and low-fitness waves (solutions) \cite{Zheng15COR}.}
\label{fig:wave}\end{figure}

The propagation updates each solution $\mathbf{x}$ by performing $\lambda(\mathbf{x})$ steps of neighborhood search, i.e., propagates $\mathbf{x}$ to a $\lambda(\mathbf{x})$-step neighboring solution. In this way, a solution with higher fitness (and therefore smaller wavelength) exploits a smaller area, while a solution with lower fitness explores a larger area, as illustrated in Fig. \ref{fig:wave}. If the resulting $\lambda(\mathbf{x})$-step neighbor is better than $\mathbf{x}$, it replaces $\mathbf{x}$ in the population.

The refraction updates a stagnant solution $\mathbf{x}$ that has not been improved for $\widehat{g}$ consecutive generations (where $\widehat{g}$ is a control parameter) by making it learn from the current best solution $\mathbf{x}^*$. At each dimension $j$, the pair of components $(x_j,x'_j)$ has a probability of 0.5 of being replaced by the corresponding components $(x^*_j,{x^*}'_j)$ of $\mathbf{x}^*$.

The breaking of a newly found best solution $\mathbf{x}^*$ generates at most $N$ one-step neighboring solutions, each being obtained by trying to replace $x'_j$ with the index of a better alternative drug in $\Lambda_{x_j}$ ($1\!\le\!j\!\le\!N$); if the best neighbor is better than $\mathbf{x}^*$, it replaces $\mathbf{x}^*$ in the population.

Algorithm \ref{alg:wwo} presents the pseudocode the WWO algorithm for the prevention program optimization problem.

\begin{algorithm}\small
Randomly initialize a population of solutions to the problem\;
\While{the stopping criterion is not satisfied}{
    Sort all solutions in increasing order of fitness\;
    Let $\mathbf{x}^*$ be the best solution in the population\;
    \ForEach{solution $\mathbf{x}$ in the population}{
        Calculate $\lambda(\mathbf{x})$ according to Eq. \eqref{eq:wavelength}\;
        \footnotesize{\tcp{propagation}}
        Let $\mathbf{x}_\lambda\!=\!\mathbf{x}$\;
        \For{$k=1$ to $\lambda(\mathbf{x})$}{
            Set $\mathbf{x}_\lambda$ to an immediate neighbor of $\mathbf{x}_\lambda$;
        }
        \If{$\textit{fit}(\mathbf{x}_\lambda)>\textit{fit}(\mathbf{x})$}{
            $\mathbf{x}\gets \mathbf{x}'_\lambda$\;
            \If{$\textit{fit}(\mathbf{x})>\textit{fit}(\mathbf{x}^*)$}{
                \footnotesize{\tcp{breaking}}
                $\mathbf{x}^*\gets \mathbf{x}$\;
                \For{$j=1$ to $N$}{
                    \If{$x'_j>1$}{
                        $x'_j\gets \textit{rand}(1,x_j\!-\!1)$\;
                        \If{$\textit{fit}(\mathbf{x})>\textit{fit}(\mathbf{x}^*)$}{
                            $\mathbf{x}^*\gets \mathbf{x}$;
                        }
                    }
                }
            }
        }
        \Else{
            \If{$\mathbf{x}$ has not been improved for $\widehat{g}$ generations}{
                Refract $\mathbf{x}$ by learning from $\mathbf{x}^*$\;
            }
        }
    }
}
\Return the best known solution $\mathbf{x}^*$.
\caption{The WWO algorithm for the prevention program optimization problem.}
\label{alg:wwo}\end{algorithm}

\section{Computational Results}\label{sec:exp}
During February and March, 2020, we have applied the proposed method to TCM prevention of COVID-19 in two regions in Zhejiang Province, China:
\begin{itemize}
\item 39,720 residents in eight communities in Hangzhou city;
\item 9,812 residents in four communities in Shaoxing city.
\end{itemize}

The following subsections report the results of fuzzy clustering of residents, TCM prevention program optimization, and prevention program implementation.

\subsection{Results of Resident Clustering}
Based on the analysis of TCM experts on local populations and TCM symptoms of COVID-19, the number $C$ of clusters is set to 16. We compare the clustering results of the standard FCM method, intuitionistic FCM (IFCM), Pythagorean FCM (PFCM), PFCM enhanced by EBO, and PFCM enhanced by the following popular metaheuristic optimization algorithms:
\begin{itemize}
\item The genetic algorithm (GA) \cite{Muhl93EC};
\item The differential evolution (DE) algorithm \cite{Storn97JGO};
\item The comprehensive learning particle swarm optimization (CLPSO) algorithm \cite{Liang06TEC};
\item The hybrid biogeography-based optimization (HBBO) algorithm \cite{Ma14EAAI}.
\end{itemize}

\begin{figure*}[!t]
\setcounter{subfigure}{0}
\begin{minipage}{0.5\linewidth}
\subfigure[Hangzhou]{\includegraphics[scale=0.42]{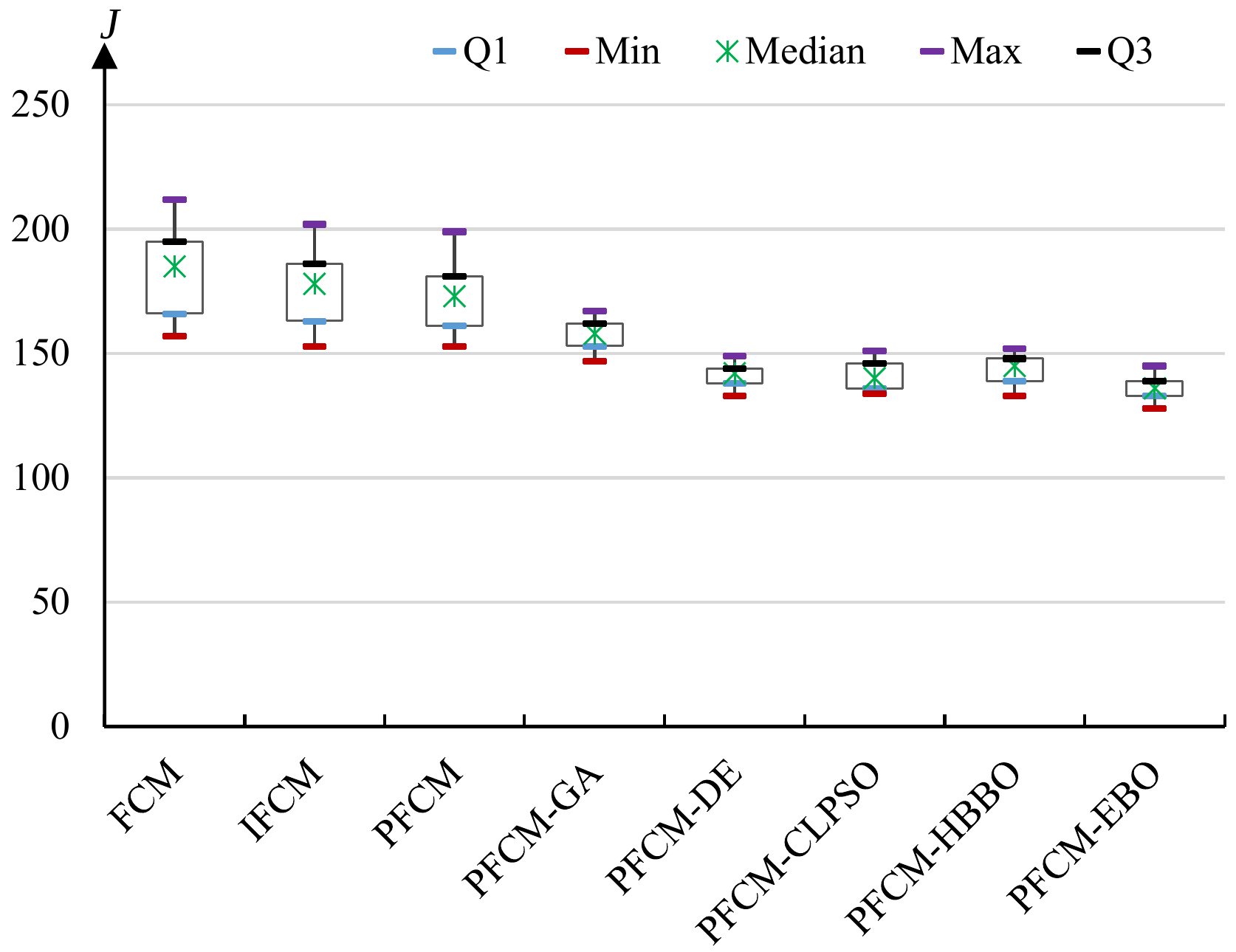}}
\end{minipage}
\begin{minipage}{0.5\linewidth}
\subfigure[Shaoxing]{\includegraphics[scale=0.42]{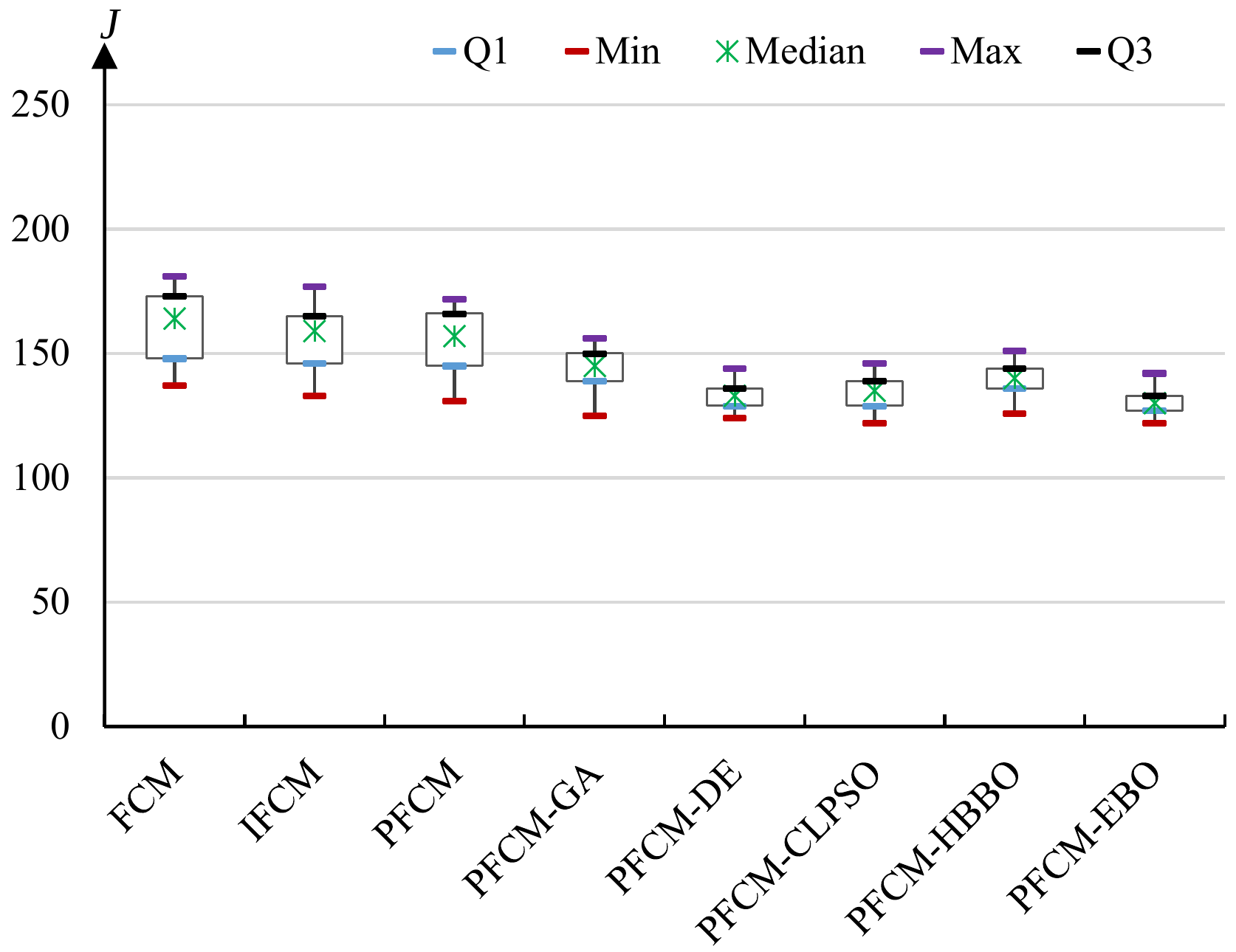}}
\end{minipage}
\hfil
\caption{Comparison of the resulting $J(U,V)$ values obtained by the algorithms for clustering residents in two cities. Each box plot shows the maximum, minimum, median, first quartile (Q1), and third quartile (Q3) of resulting $J(U,V)$ values over the 30 runs of an algorithm.}
\label{fig:res-fcm}\end{figure*}

Each algorithm runs 30 times with different random seeds (the four basic FCM methods use randomly initial cluster centroids, and the PFCM with different metaheuristics use randomly initial populations). Fig. \ref{fig:res-fcm} presents the resulting $J(U,V)$ values obtained by the eight clustering algorithm over the 30 runs. On both the instances of two regions, IFCM achieves better results than the standard FCM, and PFCM achieves better results than IFCM, which shows that using extended fuzzy sets can improve the fuzzy clustering results by capturing uncertainty information more effectively; compared to the three basic FCM methods using random initial cluster centroids, PFCM enhanced by metaheuristic optimization to find optimal/sub-optimal initial centroids achieve significant performance advantages, because the quality of initial centroids heavily affects the clustering results; among the five metaheuristic algorithm, the proposed PFCM-EBO exhibits the best performance, which demonstrates the efficiency of the EBO algorithm in optimizing initial clustering centroids for residents grouping.

\subsection{Results of Prevention Program Optimization}
The TCM experts develop 15 prevention programs for the 16 clusters of residents (two clusters are very similar and share one program). Among the 39,720 residents in Hangzhou, 4,625 residents agree to adopt the prevention programs, but medical resources required to implement the programs significantly exceed the available resource. Therefore, we use the proposed WWO algorithm to optimize the programs. Among the 15 updated programs, 13 programs are approved by the experts, and the remaining two programs are slightly modified by the experts. The resources required to implement the updated programs do not exceed the available resources and, therefore, the programs are put into implementation. However, after one week of implementation, many resources have been consumed, and the available resources are not sufficient to implement the 15 updated programs. Therefore, we perform a second round of program optimization. Among the 15 programs updated in the second round, 12 programs are approved, and the remaining three programs are modified by the experts. Nevertheless, the resources required to implement all programs exceed the available resources again. Therefore, we perform a third round of program optimization, the results of which are approved and put into implementation.

Among the 9,812 residents in Shaoxing, 1,227 agree to adopt the prevention programs. In Shaoxing, we perform two rounds of program optimization, the results of which are put into implementation successively.

In summary, we use the proposed WWO algorithm to solve five instances of the prevention program optimization problem. For comparison, we also implement the following five popular metaheuristic optimization algorithms on the five instances:
\begin{itemize}
\item The GA for constrained optimization \cite{Koziel99EC};
\item The biogeography-based optimization (BBO) for constrained optimization \cite{Ma11EAAI};
\item The DE algorithm for constrained optimization \cite{Luchi15ProcCompSci};
\item The cuckoo search (CS) algorithm for integer programming \cite{Adbel18IJCSM};
\item The grey wolf optimization (GWO) algorithm for integer programming \cite{Xing19ASOC}.
\end{itemize}

\begin{figure*}[!t]
\setcounter{subfigure}{0}
\begin{minipage}{0.33\linewidth}
\subfigure[Hangzhou, 1st round]{\includegraphics[scale=0.33]{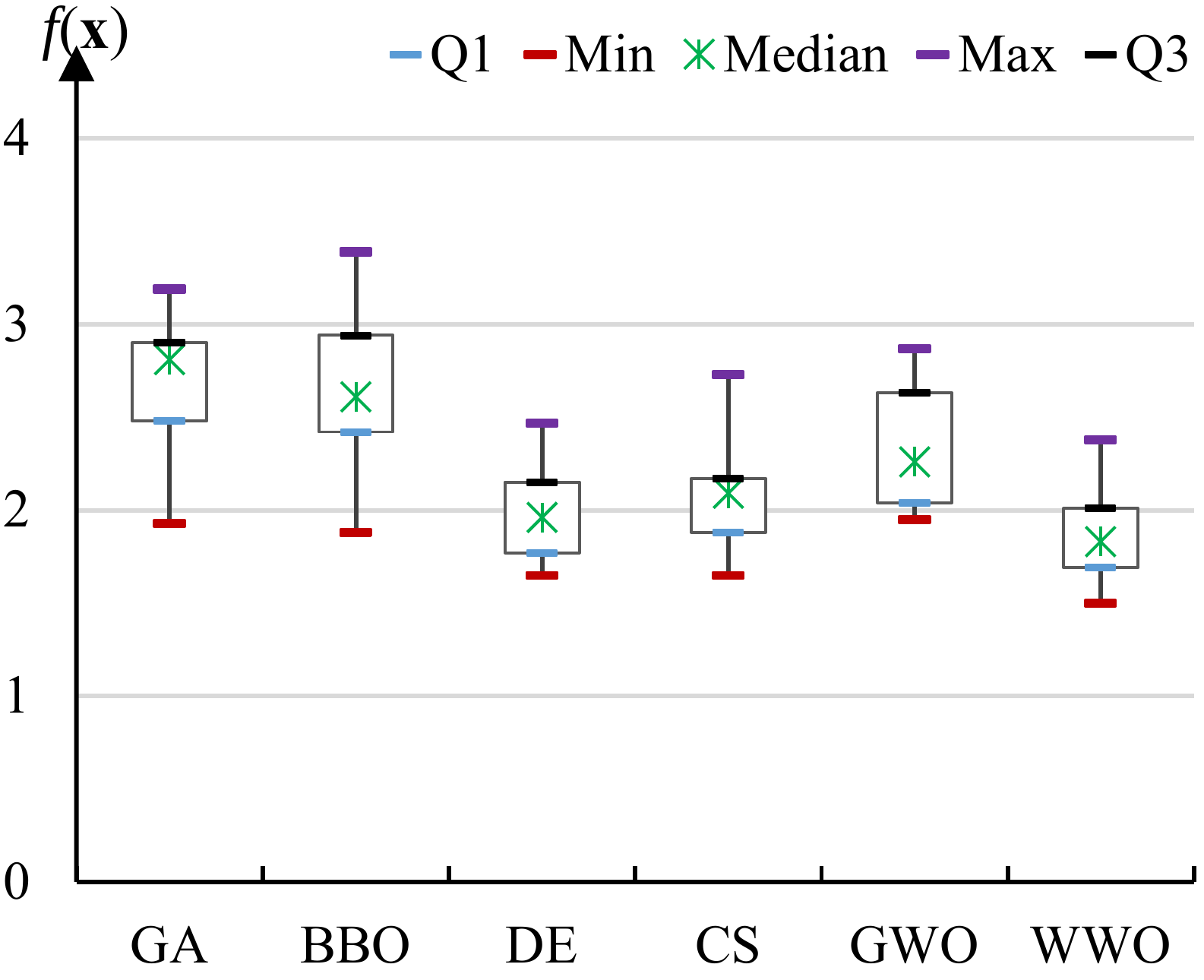}}
\end{minipage}
\begin{minipage}{0.33\linewidth}
\subfigure[Hangzhou, 2nd round]{\includegraphics[scale=0.33]{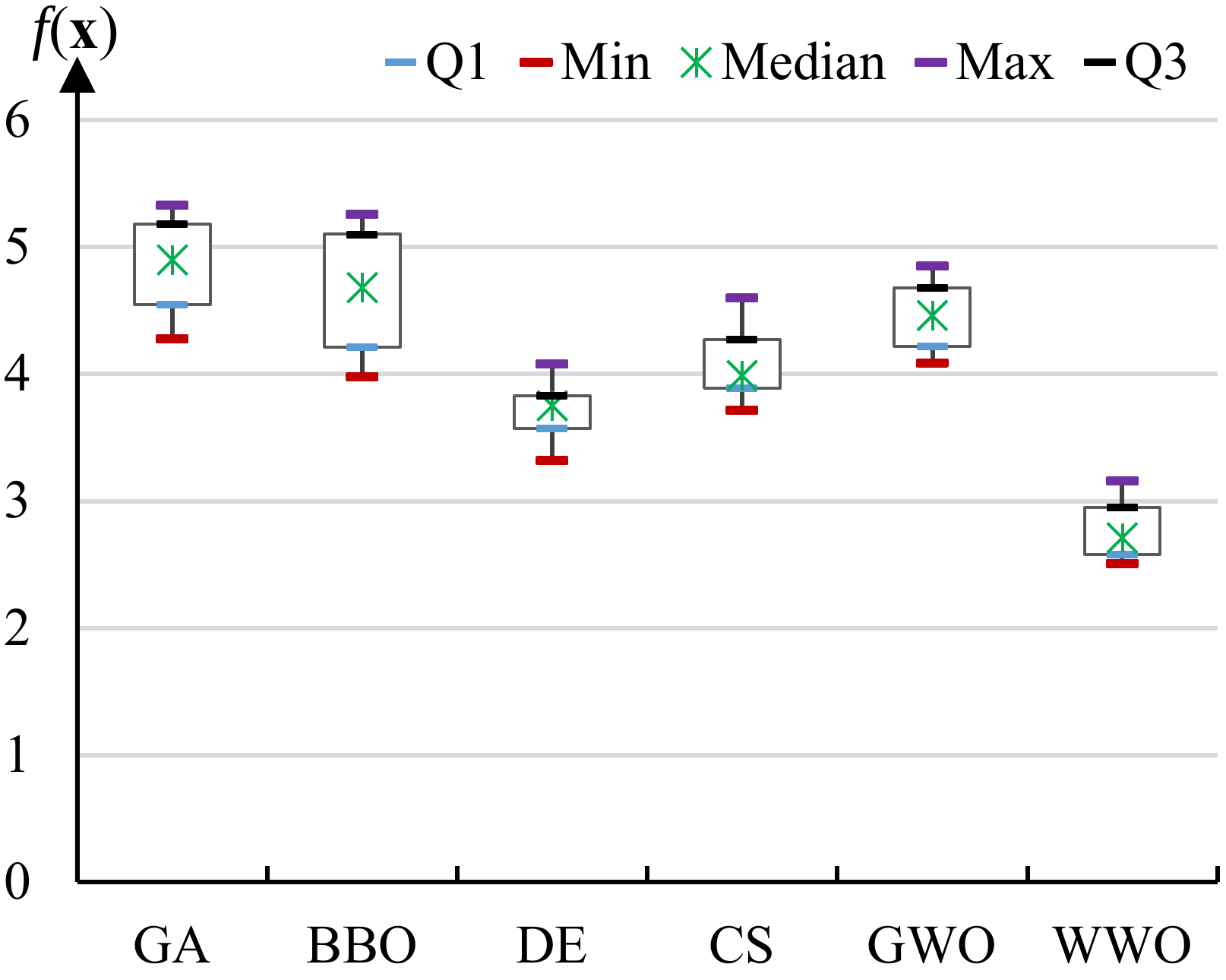}}
\end{minipage}
\begin{minipage}{0.33\linewidth}
\subfigure[Hangzhou, 3rd round]{\includegraphics[scale=0.33]{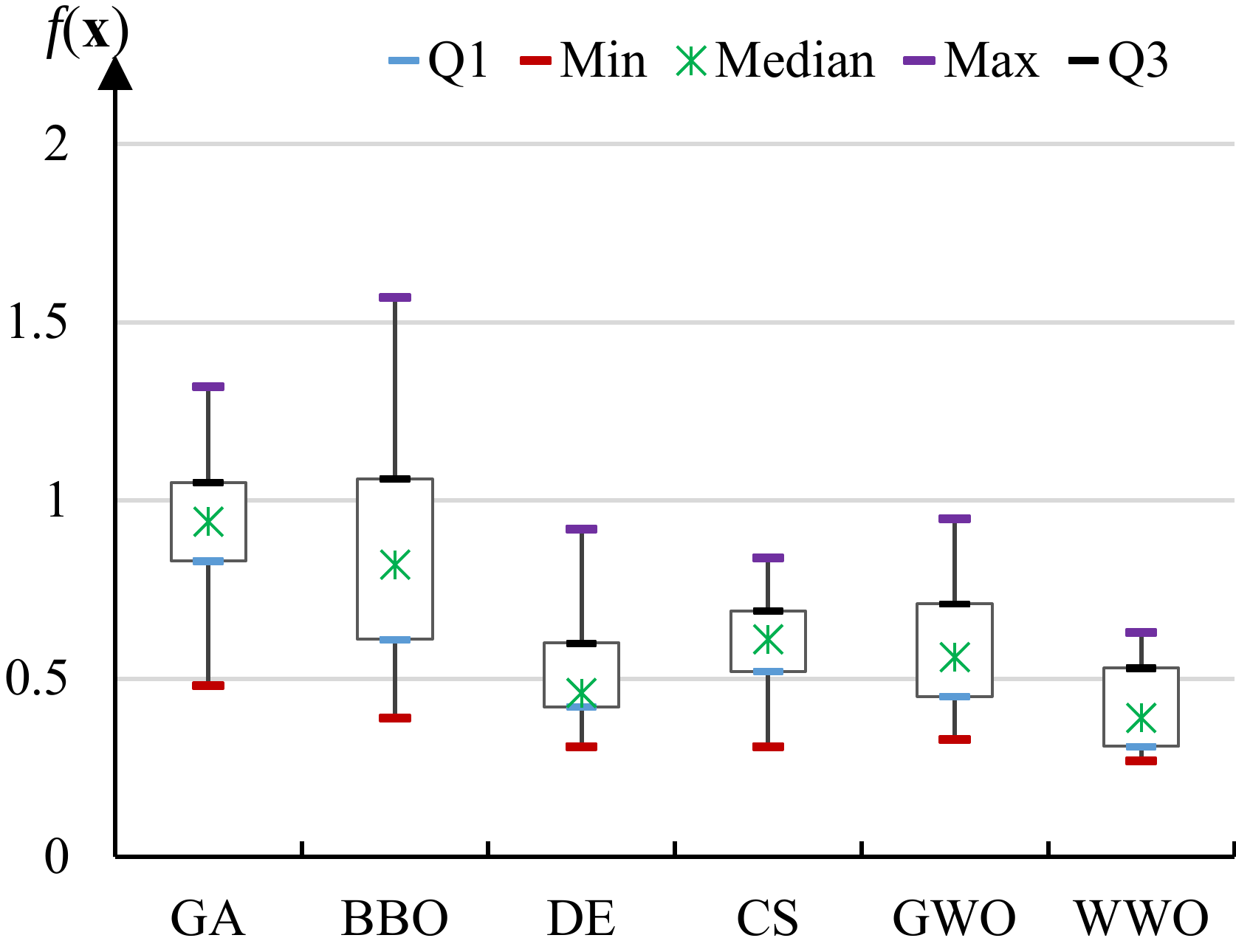}}
\end{minipage}
\begin{minipage}{0.33\linewidth}
\subfigure[Shaoxing, 1st round]{\includegraphics[scale=0.33]{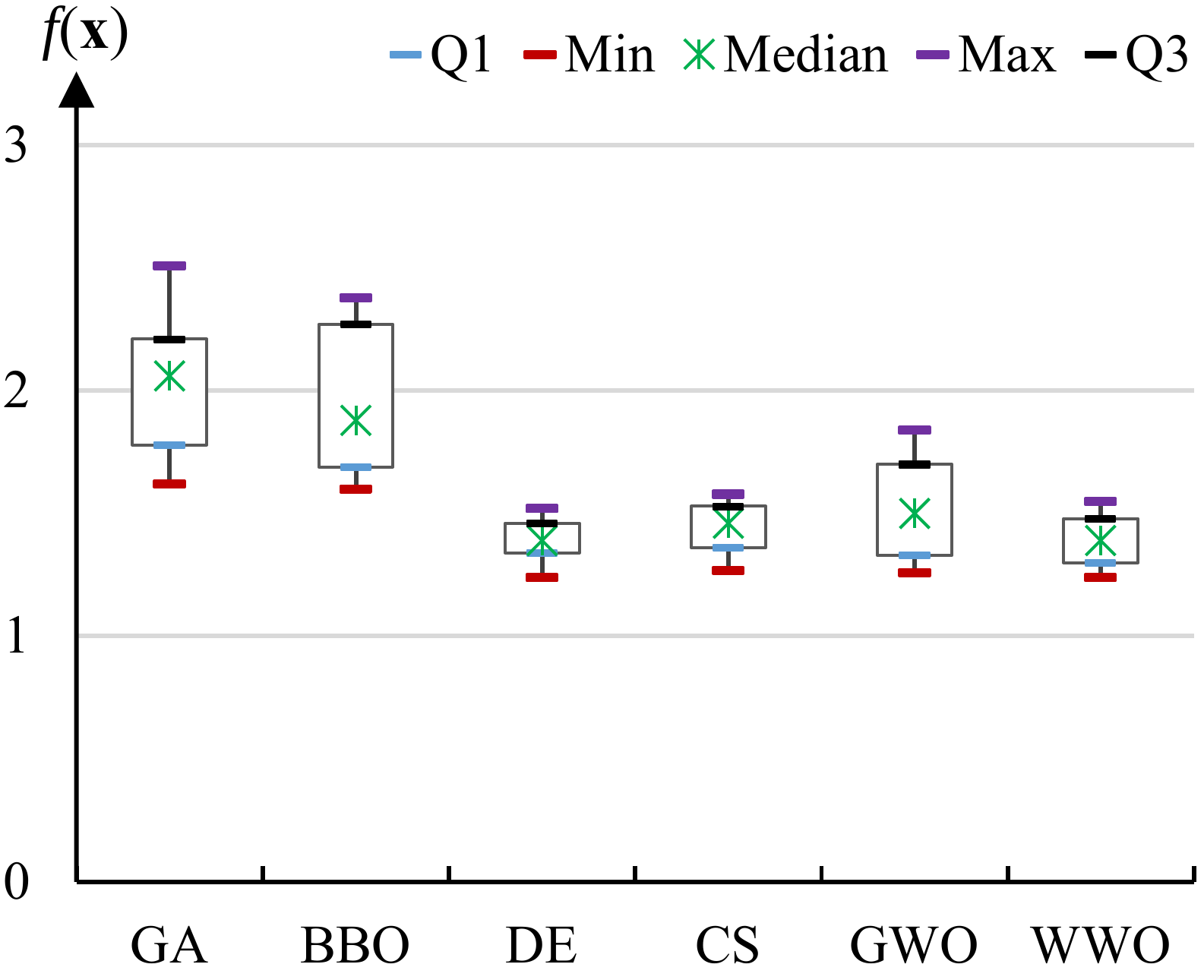}}
\end{minipage}
\begin{minipage}{0.33\linewidth}
\subfigure[Shaoxing, 2nd round]{\includegraphics[scale=0.33]{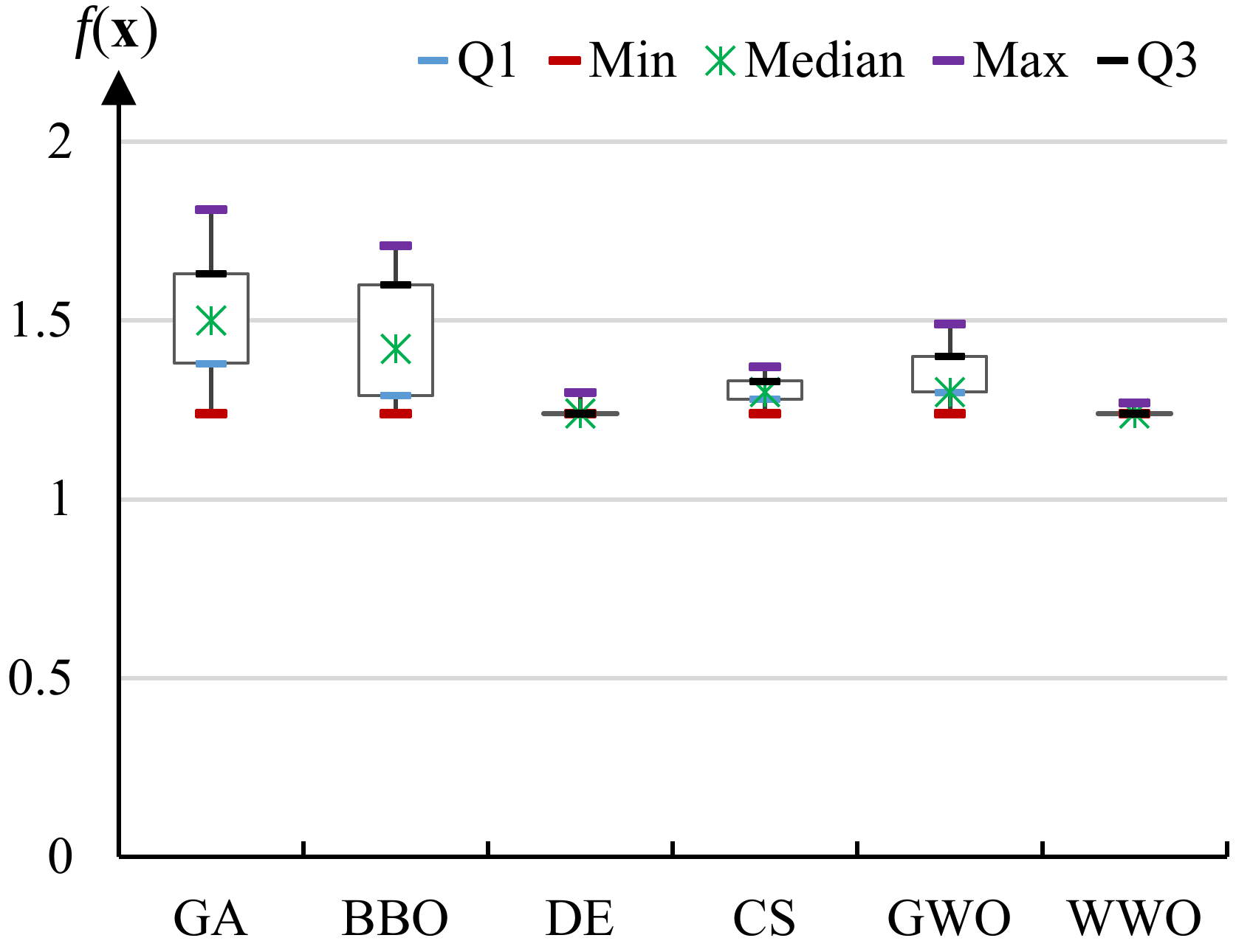}}
\end{minipage}%\hfil
\caption{Comparison of the resulting objective function values obtained by the algorithms for prevention program optimization. Each box plot shows the maximum, minimum, median, first quartile (Q1), and third quartile (Q3) of objective function values over the 30 runs of an algorithm.}
\label{fig:res-ppo}\end{figure*}

Each algorithm runs 30 times with different random seeds. Fig. \ref{fig:res-ppo} presents the resulting objective function values obtained by the six algorithm over the 30 runs. As the weights in the objective function \eqref{eq:obj} are normalized, the objective function value represents the average index of the selected alternative drugs in the alternative sets. On the three instances in Hangzhou, the median objective function values of the WWO algorithm are 1.83, 2.71, and 0.39, respectively, which are always the smallest among than the six algorithms. These three instances have the same number $N$ of programs and similar numbers of residents, but the constraints of the second-round instance is the most rigorous, while the constraints of the third-round instance is the least rigorous. The differences among the comparative algorithms are the largest on the second-round instance and the smallest on the third-round instance. On the second-round instance, the performance advantages of WWO over the other algorithms are also the most significant. This demonstrates that the proposed WWO algorithm is efficient in solving complex instances of the prevention program optimization problem.

On the two instances in Shaoxing, the median objective function values of the WWO algorithm are 1.39 and 1.24, respectively, which are also the smallest among than the six algorithms. In the second-round instance in Shaoxing, the resources required to implement the basic prevention programs do not exceed too much the available resources; on this relatively simple instance, all algorithms achieve the same minimum objective function value of 1.24, which has been verified to be the exact optimal objective function value. However, only the median objective function values of DE and WWO are 1.24, and the maximum objective function value of WWO is less than that of DE. In summary, the results show that the proposed WWO algorithm exhibits the best performance on all instances.

\subsection{Results of Prevention Program Implementation}
We compare the actual prevention effects of the diversified TCM prevention programs developed using our method with those of the uniformed TCM prevention program released by Zhejiang Provincial Health Commission. The results are presented in Table \ref{tab:res}. During February and March, 2019, in Hangzhou city, among 4,625 residents adopting our diversified prevention programs, there is no case of COVID-19 reported. During the same period, according to statistics from 71 communities in Hangzhou, among 36,138 residents adopting the uniformed prevention program, there are six cases, including five imported cases and one local case.

In Shaoxing, among 1,227 residents adopting our diversified prevention programs, there is also no case of COVID-19 reported. During the same period, among 10,530 residents adopting the uniformed prevention program, there are two imported cases.

According to the comparison results in Table \ref{tab:res}, in terms of incidence rate, the effects of diversified TCM prevention programs are obviously better than those of the uniformed TCM prevention program in both the regions.

\begin{table*}
\center\footnotesize%\setlength\tabcolsep{5pt}\renewcommand{\arraystretch}{0.9}
\caption{Comparison of the effects of our diversified TCM prevention programs with those of the uniformed TCM prevention program.}
\begin{tabular}{cc|ccccc}\hline
 &&Residents &Cases &Incidence rate &Local cases &Local incidence rate\\\hline
\multirow{2}{1.2cm}{Hangzhou} &Diversified programs &4,625 &0 &\textbf{0} &0 &\textbf{0} \\
 &Uniformed program &36,138 &6 &0.0166\% &1 &0.00277\%\\\hline
\multirow{2}{1.2cm}{Shaoxing} &Diversified programs &1,227 &0 &\textbf{0} &0 &\textbf{0}\\
 &Uniformed program &10,530 &2 &0.0190\% &0 &0\\\hline
\end{tabular}
\label{tab:res}\end{table*}

Nevertheless, due to the low incidence rate of COVID-19 in China and the limited number of residents in this study, the comparison of incidence rates does not have sufficient statistical significance. Therefore, we also conduct a questionnaire survey on the effects of TCM prevention programs. There are two questions, the first is about ``TCM prevention program helps me improve health conditions'', and the second is about ``TCM prevention program helps me prevent against COVID-19''. The answer to each question has seven choices: strongly agree, agree, weakly agree, neutral, weakly disagree, disagree, and strongly disagree.

\begin{figure*}[!t]
\setcounter{subfigure}{0}
\begin{minipage}{0.5\linewidth}
\centering\subfigure[Participants adopting diversified prevention programs]{\includegraphics[scale=0.42]{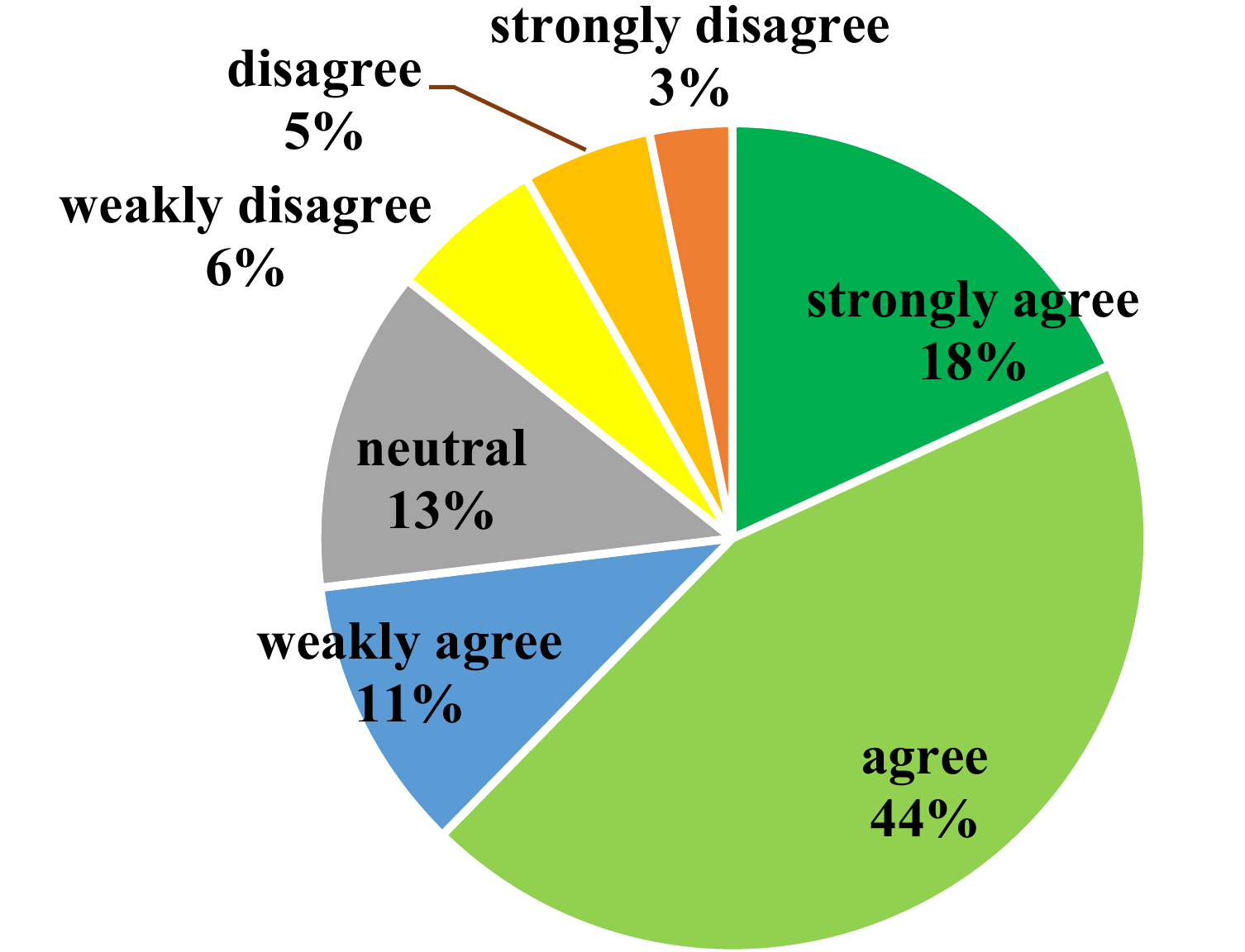}}
\end{minipage}
\begin{minipage}{0.5\linewidth}
\centering\subfigure[Participants adopting the uniform prevention program]{\includegraphics[scale=0.42]{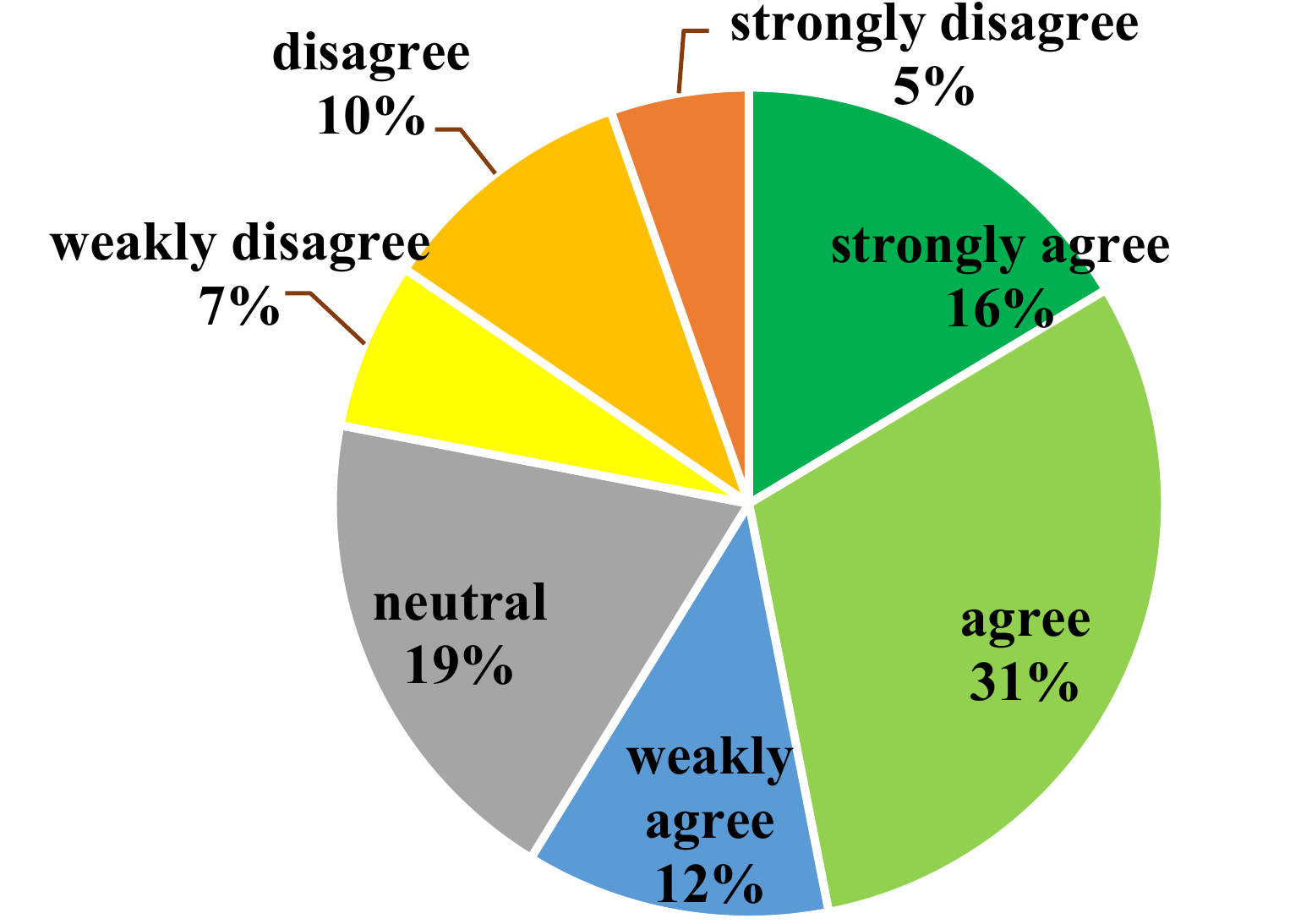}}
\end{minipage}
\caption{Distribution of different answers to the question ``TCM prevention program helps to improve health conditions''.}
\label{fig:res-stat}\end{figure*}

There are a total of 7,358 residents, including 2,550 adopting our diversified TCM prevention programs and 4,808 adopting the uniformed diversified TCM prevention program, participate the survey. Fig. \ref{fig:res-stat} presents the survey results of the question ``TCM prevention program helps me improve health conditions''. Among the participants adopting diversified prevention programs, 73\% give positive answers (18\% strongly agree, 44\% agree, and 11\% weakly agree), which is significantly higher than 59\% among the participants adopting the uniform prevention program who give positive answers (16\% strongly agree, 31\% agree, and 12\% weakly agree).

\begin{figure*}[!t]
\setcounter{subfigure}{0}
\begin{minipage}{0.5\linewidth}
\centering\subfigure[Participants adopting diversified prevention programs]{\includegraphics[scale=0.32]{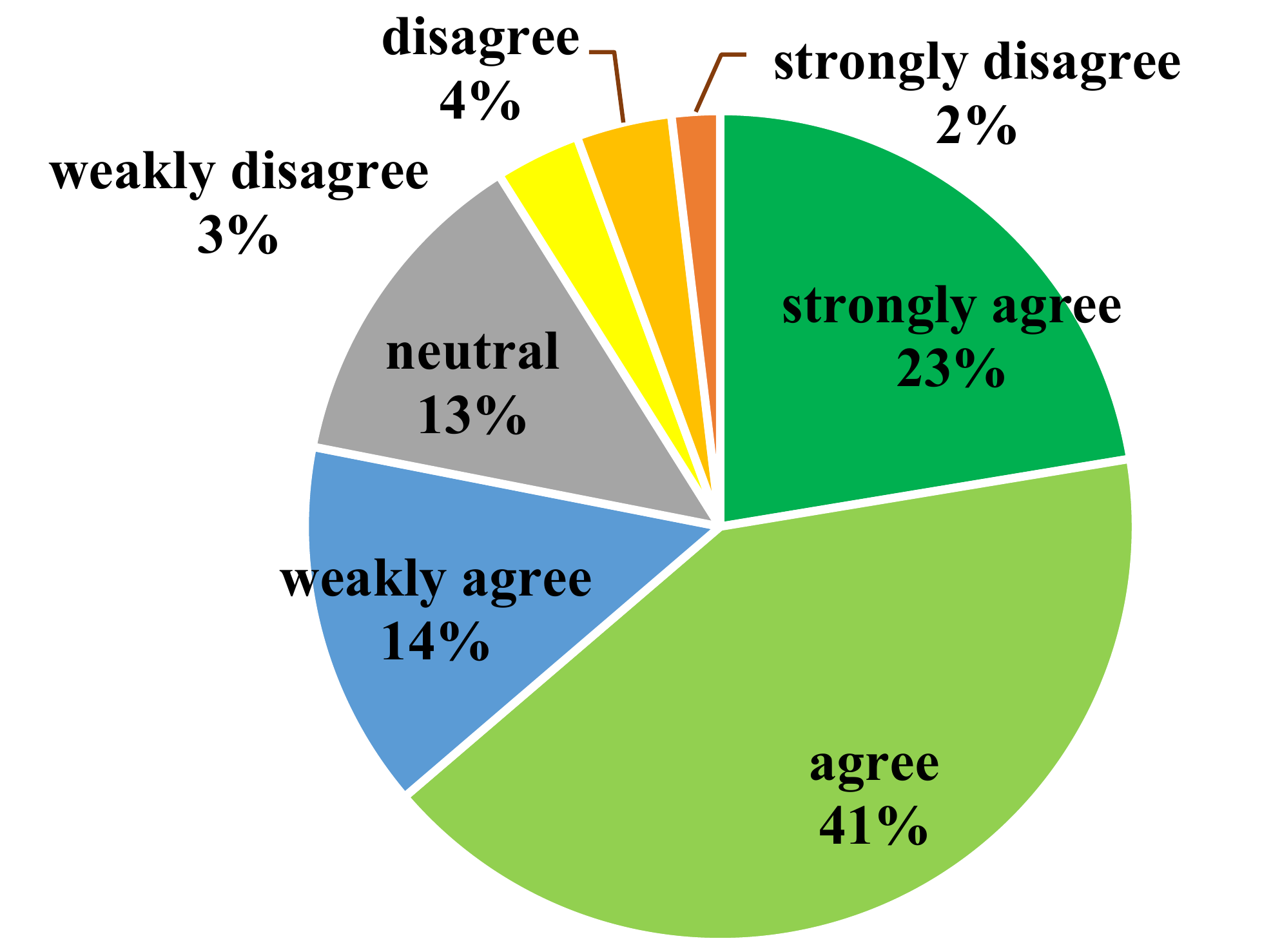}}
\end{minipage}
\begin{minipage}{0.5\linewidth}
\centering\subfigure[Participants adopting the uniform prevention program]{\includegraphics[scale=0.32]{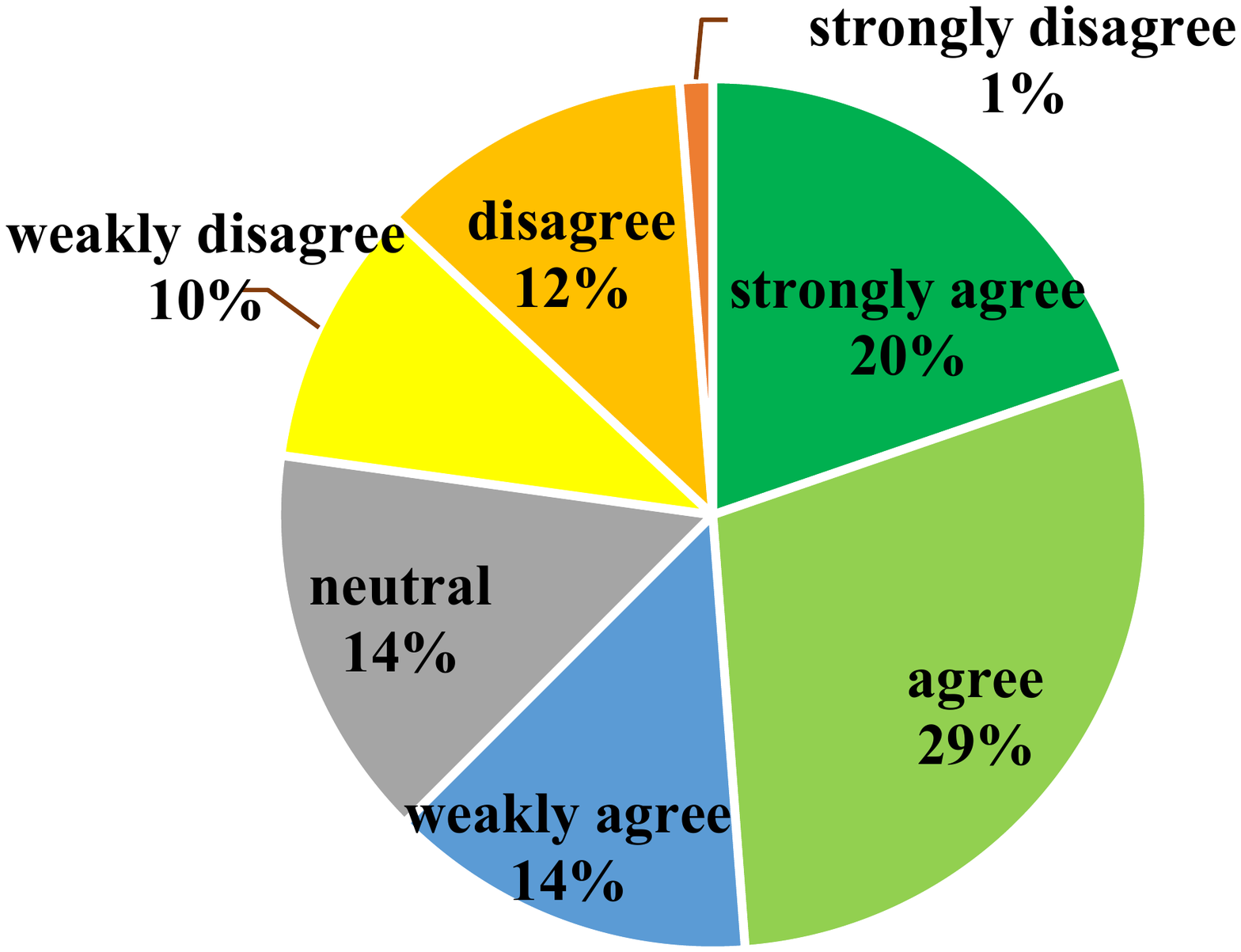}}
\end{minipage}
\caption{Distribution of different answers to the question ``TCM prevention program helps to prevent against COVID-19''.}
\label{fig:res-stat1}\end{figure*}

Fig. \ref{fig:res-stat1} presents the survey results of the question ``TCM prevention program helps me prevent against COVID-19''. Among the participants adopting diversified prevention programs, 78\% give positive answers (23\% strongly agree, 41\% agree, and 14\% weakly agree), which is significantly higher than 63\% among the participants adopting the uniform prevention program who give positive answers (20\% strongly agree, 29\% agree, and 14\% weakly agree).

In summary, the survey results demonstrate that the residents adopting diversified prevention programs are more satisfied with the effects of COVID-19 prevention and health condition improvement than the residents adopting the uniform prevention program.

\section{Conclusion}\label{sec:conclu}
In this study, we propose an intelligent optimization method to develop diversified TCM prevention programs for reducing the spread risk of and protecting populations against COVID-19. First, we use a fuzzy clustering method to divide the population based on both modern medicine and TCM health characteristics. Based on the clustering results, TCM experts develop diversified prevention programs, which are then evolved by an interactive optimization method until all the resource constraints are satisfied. The proposed method has been successfully practiced in a number of communities in Zhejiang province, China, during the peak of COVID-19.

The reported work is an emergency study aiming at COVID-19. We are continuously improving the propose method in several aspects, including (1) using health big-data analytics to enhance the feature set for clustering; (2) incorporating more TCM knowledge to the interactive optimization method to reduce the efforts of TCM experts; (3) incorporating more medical knowledge and health assessment methods to evaluate the effects of TCM prevention programs in a more accurate way. It is expected that the proposed method can be extended to the prevention and control of more epidemics in the future.

\section*{Acknowledgment}
This work was supported by National Natural Science Foundation of China under Grant No. 61872123, Zhejiang Provincial Natural Science Foundation under Grant LR20F030002 and LQY20F030001, and Zhejiang Provincial Emergency Project for Prevention \& Treatment of New Coronavirus Pneumonia under Grant 2020C03126.

%\bibliographystyle{IEEEtran}
%\bibliography{D:/paper/EnvHealth,D:/paper/bim,D:/paper/opre}
\input{ref.bbl}

\vfill
\end{document}